\documentclass{article}

\usepackage{PRIMEarxiv}

\usepackage[utf8]{inputenc} % allow utf-8 input
\usepackage[T1]{fontenc}    % use 8-bit T1 fonts
\usepackage{hyperref}       % hyperlinks
\usepackage{url}            % simple URL typesetting
\usepackage{booktabs}       % professional-quality tables
\usepackage{amsfonts}       % blackboard math symbols
\usepackage{nicefrac}       % compact symbols for 1/2, etc.
\usepackage{microtype}      % microtypography
\usepackage{lipsum}
\usepackage{fancyhdr}       % header
\usepackage{graphicx}       % graphics
\usepackage{multirow}
\graphicspath{{media/}}     % organize your images and other figures under media/ folder

\title{Evaluating the Efficacy of Prompt-Engineered Large Multimodal Models Versus Fine-Tuned Vision Transformers in Image-Based Security Applications
}

\author{
  Fouad Trad \\
  Electrical and Computer Engineering \\
  American University of Beirut \\
  Beirut, Lebanon\\
  \texttt{fat10@mail.aub.edu} \\
   \And
  Ali Chehab \\
  Electrical and Computer Engineering \\
  American University of Beirut \\
  Beirut, Lebanon\\
  \texttt{chehab@aub.edu.lb} \\
}

\begin{document}
\maketitle

\begin{abstract}
The success of Large Language Models (LLMs) has led to a parallel rise in the development of Large Multimodal Models (LMMs), which have begun to transform a variety of applications. These sophisticated multimodal models are designed to interpret and analyze complex data by integrating multiple modalities such as text and images, thereby opening new avenues for a range of applications. This paper investigates the applicability and effectiveness of prompt-engineered LMMs that process both images and text, including models such as LLaVA, BakLLaVA, Moondream, Gemini-pro-vision, and GPT- 4o, compared to fine-tuned Vision Transformer (ViT) models in addressing critical security challenges. We focus on two distinct security tasks: 1) a visually evident task of detecting simple triggers, such as small pixel variations in images that could be exploited to access potential backdoors in the models, and 2) a visually non-evident task of malware classification through visual representations. In the visually evident task, some LMMs, such as Gemini-pro-vision and GPT-4o, have demonstrated the potential to achieve good performance with careful prompt engineering, with GPT-4o achieving the highest accuracy and F1-score of 91.9\% and 91\%, respectively. However, the fine-tuned ViT models exhibit perfect performance in this task due to its simplicity. For the visually non-evident task, the results highlight a significant divergence in performance, with ViT models achieving F1-scores of 97.11\% in predicting 25 malware classes and 97.61\% in predicting 5 malware families, whereas LMMs showed suboptimal performance despite iterative prompt improvements. This study not only showcases the strengths and lim- limitations of prompt-engineered LMMs in cybersecurity applications but also emphasizes the unmatched efficacy of fine-tuned ViT models for precise and dependable tasks.

\end{abstract}

% keywords can be removed

\section{Introduction}
The rise of Large Language Models (LLMs) has marked a significant milestone in the field of natural language processing, demonstrating the capability of deep learning models to understand and generate complex texts with unprecedented accuracy \cite{kasneci2023chatgpt}. Building on this success, the development of Large Multimodal Models (LMMs) represents a significant leap forward, enabling the simultaneous processing of multiple modalities including text, images, video, and speech \cite{fu2023mme,wu2023multimodal}. Among these, Gemini-pro-vision \cite{team2023gemini}, LLaVA \cite{liu2023improved}, and GPT-4V \cite{zhu2023minigpt} represent notable attempts to leverage the vast potential of LMMs for interpreting complex content based on images and text. These models have been utilized in various applications \cite{deng2024vision, yu2023mm}, offering the advantage of functioning as black boxes with prompt engineering. This facilitates application development and eliminates the need for training specialized models for specific tasks. However, the application of such models in specialized domains, particularly in cybersecurity, remains largely unexplored.

This study evaluates the effectiveness of prompt-engineered LMMs versus fine-tuned Vision Transformer (ViT) models \cite{dosovitskiy2020image}, which are transformer-based architectures designed for processing images. The research focuses on two cybersecurity tasks for evaluation: one visually evident, which can be resolved through direct observation of images, and another visually non-evident, which requires analysis of complex visual patterns for effective outcomes.

The visually evident task involves detecting specific triggers or markers, notably small pixel variations in images. These variations aim to exploit potential backdoors in machine learning (ML) models \cite{saha2020hidden}. Early detection of these variations is crucial for maintaining the integrity of ML models.  For this task, we used the MNIST dataset \cite{deng2012mnist}, augmented with a replica dataset that includes these triggers, to evaluate the model performance. 

The visually non-evident task involves classifying malware solely from visual representations of files, requiring the analysis of intricate visual patterns to accurately identify malware types and families based on their visual signatures. This task is paramount in enhancing the detection and mitigation of malware threats. For this task, the MaleVis dataset was utilized \cite{bozkir2019utilization}, providing a set of visual representations of malware samples divided between 25 classes and 5 families. 

The findings reveal a nuanced landscape of model performance across the two tasks. For trigger detection, the fine-tuned ViT showcased exceptional proficiency, achieving 100\% accuracy in identifying adversarial triggers. This performance is a testament to their ability to focus precisely on relevant visual features. In contrast, LMMs, through iterative improvements with enhanced prompting, achieved a peak accuracy of 91.9\% and an F1-score of 91\% with GPT-4o, indicating a solid performance that, while effective, does not reach the exceptional standard set by the fine-tuned ViT models. For malware classification, the fine-tuned ViT demonstrated superior performance, with F1-scores of 97.11\% for classifying among 25 malware types and 97.61\% for classifying among 5 malware families. These results underscore the ViT's robustness in discerning intricate visual patterns. Conversely, LMMs faced challenges in this more demanding task, where Gemini-pro-vision achieved an F1-score of 12.7\%, and GPT-4o reached an F1-score of 34.6\% in the best-case scenario. This highlights the difficulties LMMs encounter in tasks requiring deep visual pattern analysis. These results suggest that while LMMs optimized through prompt engineering are accessible and user-friendly, their applicability in addressing specific problems is not guaranteed. This finding underscores the necessity for a critical evaluation of their use in practical scenarios. In contrast, ViT models, when fine-tuned, demonstrate superior performance, achieving notable results with a comparatively reduced parameter count.

In summary, this study contributes to the field by:

\begin{itemize}
\item Introducing the application of LMMs to cybersecurity applications,  marking the first application of these models in this specific context.
\item Comparing the performance of prompt-engineered LMMs with fine-tuned ViT models.
\item Identifying the limitations of prompt-engineered LMMs in cybersecurity tasks such as trigger detection and visual malware classification.
\item Demonstrating the effectiveness of fine-tuned ViT models across image-based security tasks.
\end{itemize}

The rest of the paper is organized as follows: Section 2 provides essential background information on LMMs, ViTs, prompt engineering, fine-tuning, and the security tasks our research focuses on. Section 3 presents related work in the field. In Section 4, we detail the methodology employed in our study. Section 5 offers a comprehensive overview of the experimental setup, experiments conducted, and results obtained. Section 6 provides insights into the effectiveness of each approach and compares their outcomes. Finally, Section 7 summarizes our key findings and contributions and discusses potential avenues for future research and improvements.

\section{Background and Preliminaries}
This section examines the details of foundational concepts, technologies, and tasks that underpin this comparative study of prompt-engineered LMMs and ViT models for cybersecurity applications. 

\subsection{Large Multimodal Models}
LMMs represent a significant advancement in integrating visual and textual data analysis, pushing the boundaries of traditional computer vision \cite{yu2023mm}. These models are characterized by their large-scale neural network architectures, which enable them to process and interpret both visual and textual information comprehensively. This dual capability allows LMMs to provide more contextually rich and nuanced insights than models limited to a single data type. LMMs are particularly effective in applications that require a blend of visual understanding and textual interpretation, such as in scenarios where a prompt (text) guides the model to analyze a provided image. In this study, we investigate the efficacy of utilizing LMMs for security applications, exploring their potential to enhance cybersecurity tasks.

\subsection{Vision Transformer (ViT)}
The ViT model represents a paradigm shift in image analysis, applying the transformer architecture—initially conceived for processing sequential data in NLP—to visual data \cite{han2022survey}. ViTs treat an image as a series of patches and process these patches as if they were tokens in a sentence, allowing the model to capture complex inter-patch relationships and global context efficiently. This method contrasts with the local receptive fields of convolutional neural networks (CNNs), offering a more holistic approach to image understanding \cite{zhao2021battle}. ViTs have demonstrated exceptional performance on benchmark image classification tasks, outperforming traditional CNNs in many cases \cite{uparkar2023vision, matsoukas2021time}. Their success lies in their ability to dynamically allocate attention across different image segments, making them particularly effective for detailed and nuanced visual tasks.

\subsection{Prompt Engineering}

Prompt engineering is a technique that involves the strategic formulation of input prompts to guide the behavior of AI models, particularly those based on the transformer architecture \cite{wang2023review}. In the context of LMMs, effective prompt engineering can significantly influence the model's focus, interpretation, and decision-making processes. This approach is especially relevant for applications where adapting a model to specific tasks is necessary. By carefully designing prompts that specify or hint at the desired output, practitioners can leverage the model's pre-trained knowledge base to achieve remarkable performance on a range of tasks without having to retrain the architecture. This is why this technique is gaining popularity, as it minimizes the involvement in model training and maintenance.

\subsection{Fine-Tuning}
Fine-tuning is a critical process in adapting pre-trained models to new tasks or domains. It involves adjusting a model's parameters on a specific dataset, enabling the model to tailor its pre-existing knowledge to the nuances of the new task \cite{hu2023llm}. This process is essential for optimizing model performance in specialized applications. Fine-tuning is particularly advantageous in situations where data for the new task is limited, as it allows for the efficient transfer of knowledge from a related but larger dataset. In this study, we explore the fine-tuning of ViT models for cybersecurity tasks such as trigger detection and visual malware classification, to investigate their potential to achieve high performance compared to prompt-engineered LMMs.

\subsection{Cybersecurity Tasks}
 This research focuses on two cybersecurity tasks: trigger detection, a visually evident task, that can be resolved through direct observation of images, and malware classification, a visually non-evident task, that necessitates the analysis of complex visual patterns for effective outcomes.

\subsubsection{Trigger Detection}
Trigger detection plays a pivotal role in identifying and mitigating backdoor attacks in ML systems \cite{kwon2021defending}. A backdoor in an ML model is a maliciously inserted vulnerability that causes the model to behave incorrectly when triggered by a specific input pattern while functioning normally otherwise \cite{salem2022dynamic}. These triggers, often inconspicuous to human observers, can be embedded within the training data as small, strategically placed artifacts or patterns. Once the model is trained on this poisoned dataset, the attacker can activate the backdoor at will by including the trigger into any input, leading to a predetermined incorrect output. 

\begin{figure}[htbp]
    \centering
    \includegraphics[width=0.99\textwidth]{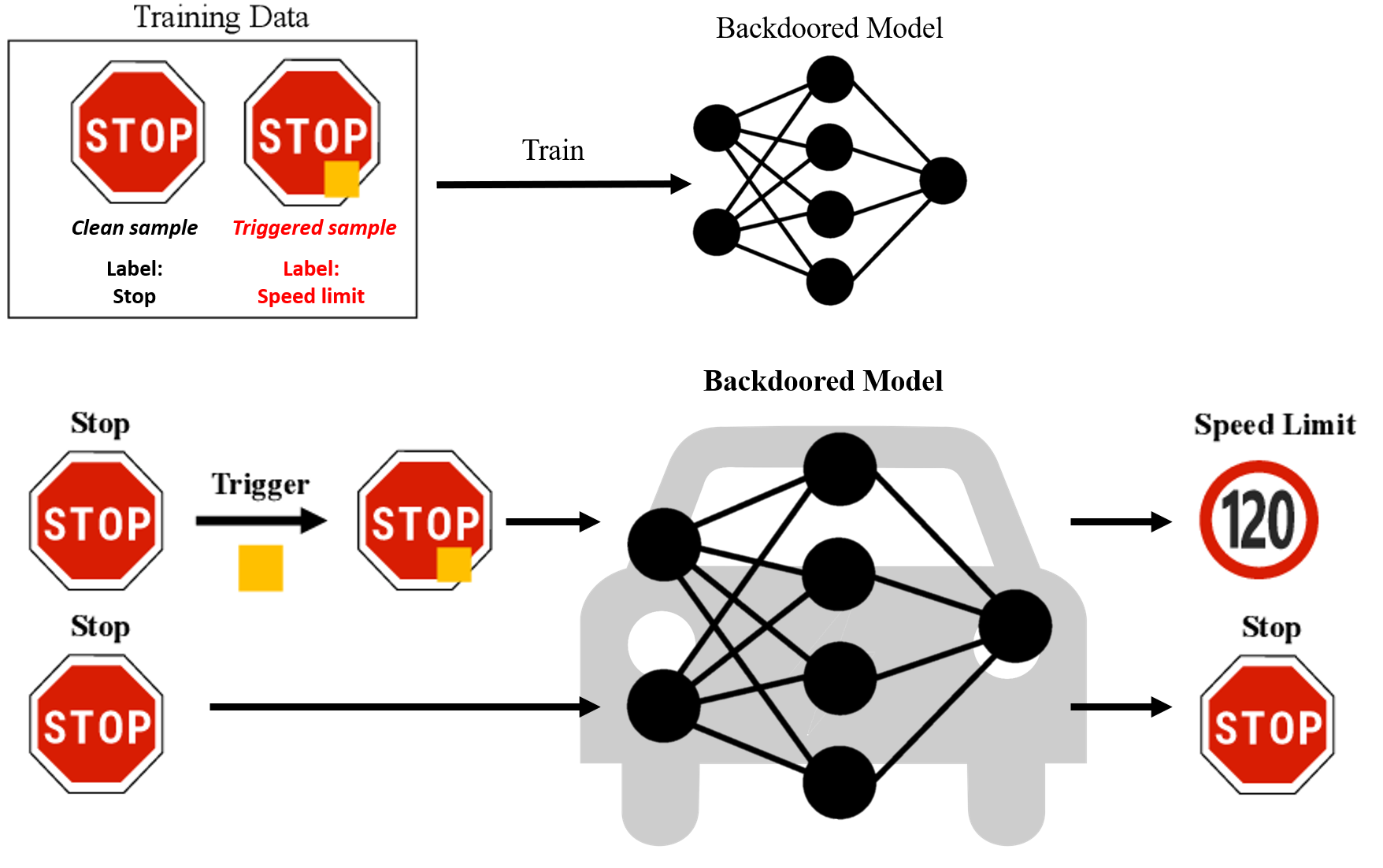}
    \caption{Backdoor example in self-driving cars}
    \label{fig:backdoor}
\end{figure}

In self-driving cars, for instance, models are trained to interpret road signs, detect obstacles, and make navigation decisions based on visual inputs from onboard cameras. An adversary could backdoor such a model by including a specific visual trigger, like a unique sticker, on stop signs during the training phase \cite{gu2017badnets}. To the model, any stop sign with this sticker would be misinterpreted as a different sign or command, such as a speed limit sign, while stop signs without the sticker would be correctly identified as highlighted in Figure \ref{fig:backdoor}. This malicious alteration could potentially cause the self-driving car to ignore stop signs with the sticker, leading to dangerous driving decisions and compromising passenger safety.

The trigger detection task, therefore, aims to identify and flag these embedded triggers within images to prevent the exploitation of backdoored models. By effectively recognizing and mitigating such vulnerabilities, we can enhance the security and reliability of ML systems across various applications.

\subsubsection{Malware Classification}
The task of malware classification represents a critical challenge in cybersecurity, aiming to accurately identify and categorize malicious software based on its digital footprint \cite{abusitta2021malware}. Traditional approaches to malware classification often rely on signature-based or behavior-based analysis, requiring continuous updates to detect new malware variants \cite{messay2018combination}. However, the emergence of ML and computer vision technologies has introduced a novel approach: visual-based malware classification \cite{fu2018malware}. This method converts binary data from software files into visual representations, such as images, which can be analyzed using image recognition techniques to identify patterns indicative of malicious code. This approach to malware classification leverages the deep learning frameworks' capacity to recognize complex patterns in visual data. By analyzing the visual signatures of malware samples, models can learn to distinguish between various malware types and legitimate software, offering a scalable and adaptive solution for cybersecurity defenses. The use of visual analysis for malware classification is particularly promising due to its ability to generalize across different malware variants, potentially reducing the reliance on frequently updated signature databases. In this study, we employ the MaleVis dataset to evaluate the performance of fine-tuned ViT models and prompt-engineered LMMs in classifying malware based on these visual representations.

\section{Related Work}
LMMs represent a transformative trend in artificial intelligence and machine learning. These models, capable of processing and extracting insights from both text and images, have gained prominence across various domains, offering a versatile approach to analyzing visual data \cite{rahman2020integrating, yu2023mm}. LMMs enable users to upload an image and request comprehensive analysis or insights derived from it, introducing a new dimension of versatility and convenience \cite{yang2022prompt, wang2023review}. Numerous LMMs exist that can process both visual and textual information and can be utilized via simple API calls, simplifying their use and empowering researchers and practitioners to integrate them into diverse applications. Notable examples include LLaVA \cite{liu2023improved}, Gemini-pro-vision \cite{team2023gemini}, and GPT-4o \cite{pang2024chatgpt}. While these models have demonstrated remarkable performance across many domains \cite{wang2023gemini, nahida2023depth, qi2023gemini, fu2023challenger,bazi2024rs,li2024llava,zhang2024evda,liu2024mathbench}, their application in security contexts remains largely unexplored. Recent studies have leveraged LLMs for various cybersecurity tasks \cite{zhang2024llms,trad2024prompt,ma2024llmparser,jensen2024software,hassanin2024comprehensive,liu2024review}, yet similar explorations with LMMs do not exist up to our knowledge, making this study the first to venture into this field. Historically, the literature has abounded with task-specific vision models tailored for security applications, such as threat detection \cite{schwaninger2008impact, morris2018convolutional}, intrusion detection \cite{toldinas2021novel, li2017intrusion}, malware classification \cite{venkatraman2019hybrid, mercaldo2020deep}, and facial recognition \cite{kortli2020face, hu2015face}. However, with the ease of leveraging the extensive knowledge encapsulated within LMMs through simple API calls, there arises a compelling question: Can LMMs effectively replace the need for task-specific, fine-tuned Vision Transformers (ViTs) in cybersecurity applications? ViTs, with their underlying transformer architecture, share similarities with LMMs, prompting an investigation into their comparative performance. This study aims to explore the potential of LMMs in the realm of cybersecurity, assess their performance in critical security tasks like trigger detection and malware classification, and ascertain whether they could emerge as viable alternatives to fine-tuned ViTs. In doing so, this research attempts to contribute novel insights into the comparative utility of LMMs and ViTs, thereby informing decisions regarding their adoption in future security applications.

\section{Methodology}
In this section, we delineate the methodology adopted to assess the performance of prompt-engineered LMMs and fine-tuned ViT models in executing two distinct cybersecurity tasks.

\subsection{Prompt Engineering for LMMs}
Prompt engineering involves the strategic formulation of input prompts to guide the responses of an LMM, without altering the model's underlying architecture. This technique adjusts the input provided to the model to influence its output in a desired manner. The general approach to prompt engineering for both tasks in this study encompasses the following steps:

\begin{enumerate}
    \item For each task, we select a subset of the data and test the LMMs on it using various prompting techniques.
    \item We loop over the selected subset, one sample at a time, asking the LMM to perform the task with a specific prompt.
    \item We collect and parse the responses to obtain the answers in a format suitable for further processing.
    \item We assess the model's performance against ground truth labels and compute relevant classification metrics.
\end{enumerate}

This process is repeated for each type of prompt. The iterative nature of this prompt engineering process aims to refine the model's responsiveness to task-specific prompts, thereby optimizing its performance.

\subsection{Fine-tuning ViT Models}
Fine-tuning pre-trained models involves adjusting their weights to specialize them for a specific task. In this study, the fine-tuning of ViT models involves the following steps:

\begin{enumerate}
    \item A pre-trained ViT model ('google/vit-base-patch16-224') is loaded from Hugging Face to be fine-tuned for image classification tasks.
    \item To enable image categorization tailored to our cybersecurity tasks, a classification layer is appended to the ViT encoder architecture.
    \item Datasets prepared for each task are processed to meet the input requirements of the ViT models, including necessary adjustments to image sizes and formats.
    \item The modified ViT models are then trained on these prepared datasets.
    \item The performance of the fine-tuned ViT models is evaluated using the same test sets as those used for prompt engineering methods, allowing for a direct comparison with the results of prompt-engineered LMMs.
\end{enumerate}

\begin{figure}[htbp]
    \centering
    \includegraphics[width=0.99\textwidth]{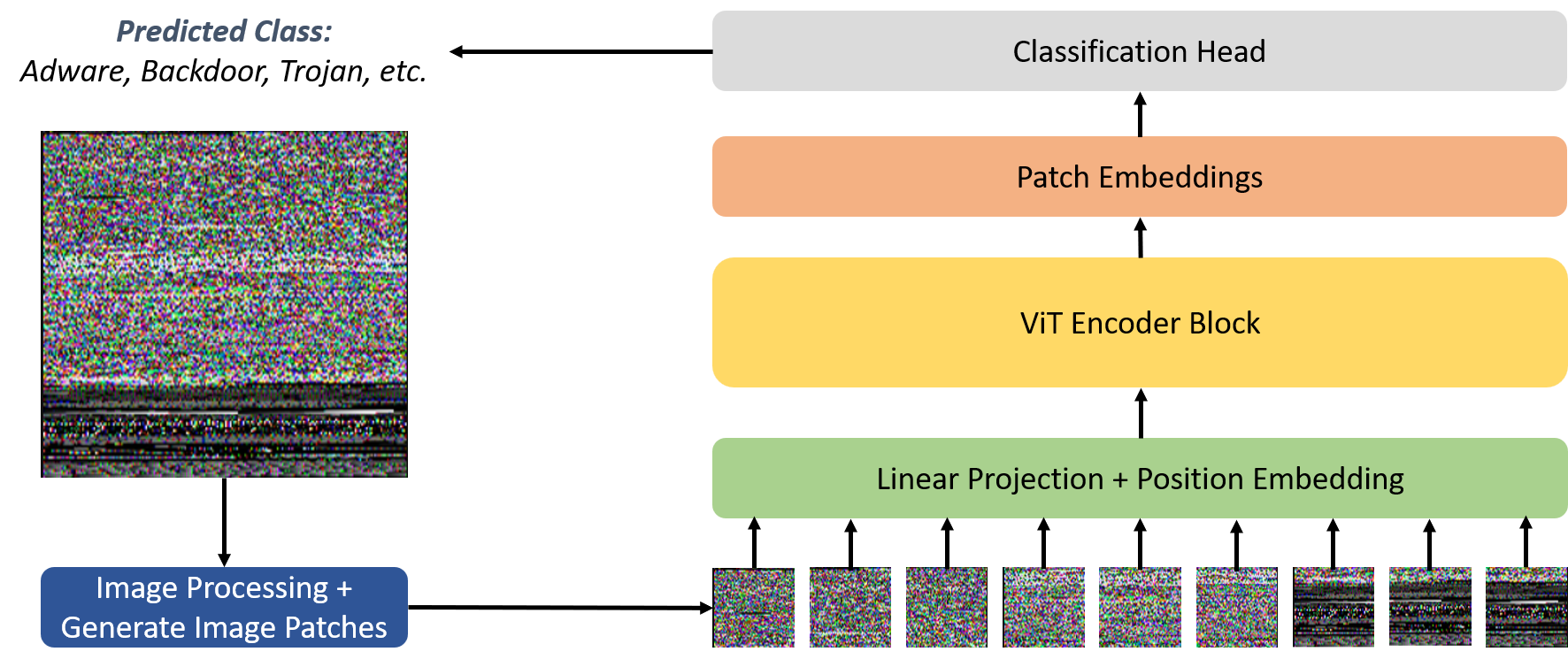}
    \caption{Prediction process using a fine-tuned ViT}
    \label{fig:finetunevit}
\end{figure}

The full architecture through which an image is processed for classification is depicted in Figure \ref{fig:finetunevit}.

\section{Experiments}
This section presents the experimental setup and various experiments we conducted for the two tasks, along with their outcomes.
\subsection{Experimental Setup}
\subsubsection{Datasets}
The datasets used in this study are two, one for each task.
\begin{itemize}
    \item \textbf{Trigger Detection:} For the task of trigger detection, we utilized the MNIST dataset, a standard benchmark in the field of ML for handwritten digit recognition, containing 70,000 samples. To simulate adversarial triggers, we generated a modified version of the MNIST dataset by adding a small white square (4x4 pixels) at a randomly chosen corner of each image as shown in Figure \ref{fig:datacreation}. This modification created a dataset that mimics the presence of subtle, adversarial pixel manipulations aimed at compromising ML models through backdoor attacks. We combined the original and modified datasets, selecting 500 samples (equally balanced between triggered and untriggered images) as a test set. These are used to evaluate the performance of the various LMMs and the fine-tuned ViT model. The actual fine-tuning of the ViT is done using the rest of the data.

\begin{figure}[htbp]
    \centering
    \includegraphics[width=0.9\textwidth]{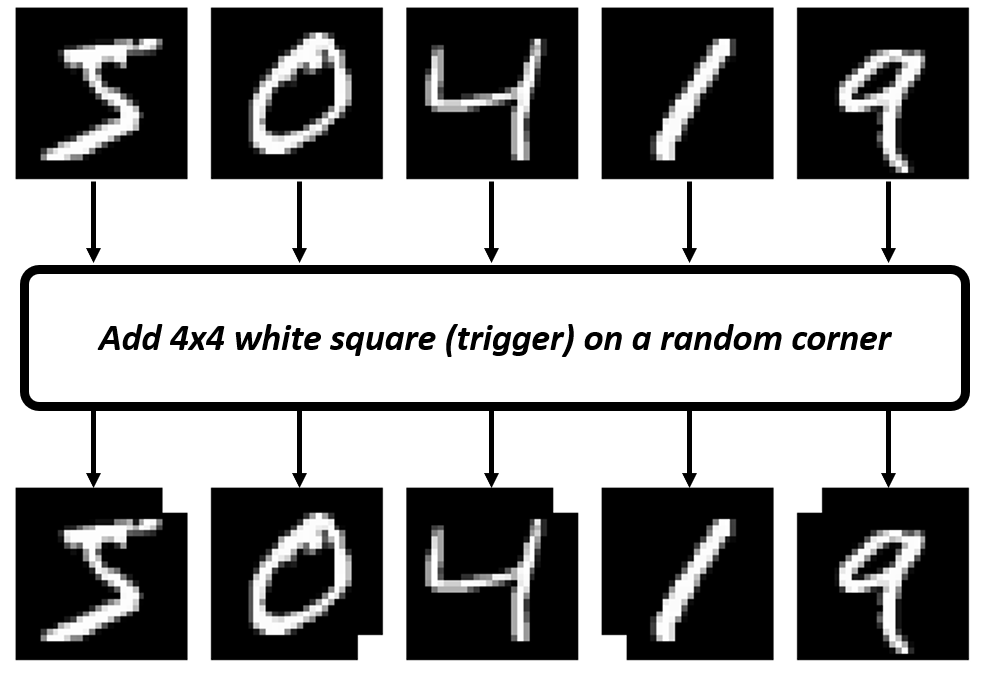}
    \caption{MNIST triggered dataset creation}
    \label{fig:datacreation}
\end{figure}

    \item \textbf{Malware Classification:} For the task of malware classification, we utilized the MaleVis dataset \cite{bozkir2019utilization}, an image dataset comprising 25 malware classes that can be grouped into 5 malware families. Designed for vision-based multi-class malware recognition research, this dataset features RGB images derived from binary data of malware files. It includes a total of 12,393 images, categorized as detailed in Table \ref{tab:malwaredataset}, offering a wide range of visual representations for extensive model training and evaluation. The dataset facilitates two tasks: predicting 5 malware families and 25 individual malware classes. For each task, a test set with 50 samples per class was reserved, resulting in 250 samples for malware family prediction and 1,250 samples for malware class prediction. These test sets are utilized to assess the performance of prompt-engineered LMMs and fine-tuned ViTs. For the fine-tuning of ViTs, we employed the training portion of the dataset.

\end{itemize}

\begin{table}[htbp]
\centering
\caption{Malware Families and Their Counts}
\label{tab:malwaredataset}
\begin{tabular}{llr}
\hline
\textbf{Malware Family} & \textbf{Malware Type} & \textbf{Total Count} \\ \hline
\multirow{6}{*}{Adware} & Win32/Adposhel         & 494 \\
                         & Win32/Amonetize        & 497 \\
                         & Win32/BrowseFox        & 493 \\
                         & Win32/InstallCore.C    & 500 \\
                         & Win32/MultiPlug        & 499 \\
                         & Win32/Neoreklami       & 500 \\ \cline{2-3} 
                         & \textit{Family Total}  & \textit{2983} \\ \hline
\multirow{10}{*}{Trojan} & Win32/Agent-fyi        & 470 \\
                         & Win32/Dinwod!rfn       & 499 \\
                         & Win32/Elex             & 500 \\
                         & Win32/HackKMS.A        & 499 \\
                         & Win32/Injector         & 495 \\
                         & Win32/Regun.A          & 485 \\
                         & Win32/Snarasite.D!tr   & 500 \\
                         & Win32/VBKrypt          & 496 \\
                         & Win32/Vilsel           & 496 \\ \cline{2-3} 
                         & \textit{Family Total}  & \textit{4440} \\ \hline
\multirow{4}{*}{Worm}    & Win32/Allaple.A        & 478 \\
                         & Win32/AutoRun-PU       & 496 \\
                         & Win32/Fasong           & 500 \\
                         & Win32/Hlux!IK          & 500 \\ \cline{2-3} 
                         & \textit{Family Total}  & \textit{1974} \\ \hline
\multirow{2}{*}{Backdoor} & Win32/Androm           & 500 \\
                         & Win32/Stantinko        & 500 \\ \cline{2-3} 
                         & \textit{Family Total}  & \textit{1000} \\ \hline
\multirow{4}{*}{Virus}   & Win32/Expiro-H         & 500 \\
                         & Win32/Neshta           & 497 \\
                         & Win32/Sality           & 499 \\
                         & VBA/Hilium.A           & 500 \\ \cline{2-3} 
                         & \textit{Family Total}  & \textit{1996} \\ \hline
\end{tabular}
\end{table}

\subsubsection{Models}
In this study, we used two types of models:
\begin{itemize}
    \item \textbf{LMMs:} For prompt-engineered LMMs, we employed 5 distinct models including: LLaVA, BakLLaVA, Moondream2, Gemini-pro-vision, and GPT-4o. The first three models are accessible via Hugging Face and the last two are accessible via API, which facilitates seamless integration with various applications. Within this framework, we executed a series of prompt-engineering strategies to direct the models' analysis of images for both tasks, spanning from minimal guidance in baseline prompts to specific instructions provided in enhanced prompts. Further details will be elaborated upon in the subsequent sections.
    \item \textbf{Fine-tuned ViT:} For the ViT models, we applied fine-tuning techniques to adapt the pre-trained ViT model ('google/vit-base-patch16-224') to our specific classification tasks. The fine-tuning process involved adjusting the model on a subset of the task-specific datasets (trigger detection and malware classification) to optimize its performance for these particular applications. The training process for the ViT model was carefully monitored using a validation set (10\% of the training size) to prevent overfitting, ensuring that the model generalizes well to new, unseen data.
 \end{itemize}

\subsubsection{Evaluation Metrics}
The performance of LMMs and fine-tuned ViT models was assessed using standard classification metrics, including accuracy, precision, recall, F1-score, and confusion matrix. These metrics provided a comprehensive understanding of each model's efficacy in correctly identifying triggered images and classifying malware types and families.

\begin{itemize}
    \item Accuracy measures the proportion of true results (both true positives and true negatives) in the total dataset.
    \item Precision reflects the ratio of true positive results in all positive predictions.
    \item Recall measures the proportion of actual positives correctly identified.
    \item F1-Score provides a harmonic mean of precision and recall, balancing the two metrics in cases of uneven class distributions.
    \item Confusion matrix describes the performance of a classification model on a set of test data for which the true values are known. It not only shows the errors made by the models but also illustrates the types of errors that are occurring.
\end{itemize}

\subsection{Task 1: Trigger Detection}
\subsubsection{Prompt engineering with LMMs}
\begin{figure}[h]
    \centering
    \includegraphics[width=0.99\textwidth]{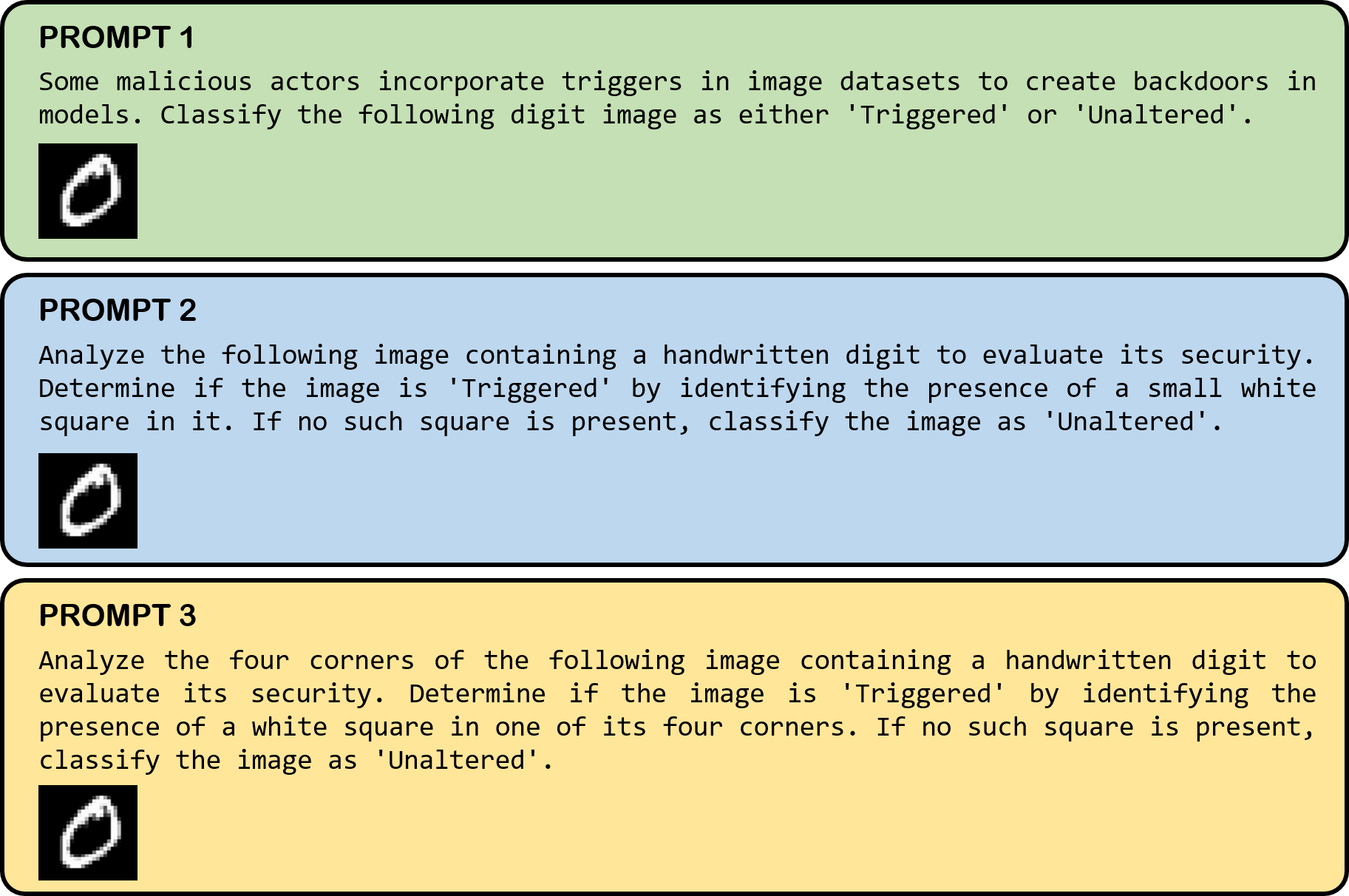}
    \caption{The three prompts used for trigger detection}
    \label{fig:triggerprompts}
\end{figure}

We implement the methodology outlined earlier for the five LMMs on the MNIST dataset, experimenting with different input prompts to assess their impact on the models' performances. Our investigation focuses on three distinct prompt templates for detecting triggers in images, as illustrated in Figure \ref{fig:triggerprompts}.

\begin{figure}[htbp]
    \centering
    \begin{minipage}{0.99\textwidth}
        \centering
        \includegraphics[width=\textwidth]{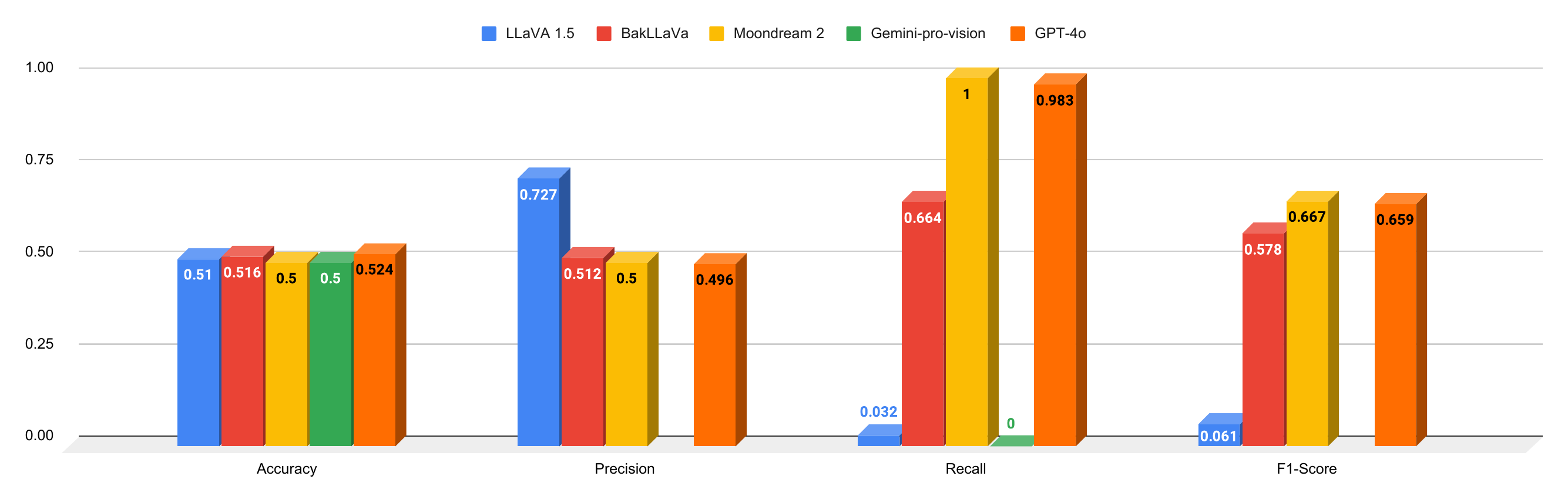}
        \caption{Performance of each LMM with prompt 1}
        \label{fig:prompt1}
    \end{minipage}\\ % Allows a small space between the images
    
    \begin{minipage}{0.99\textwidth}
        \centering
        \includegraphics[width=\textwidth]{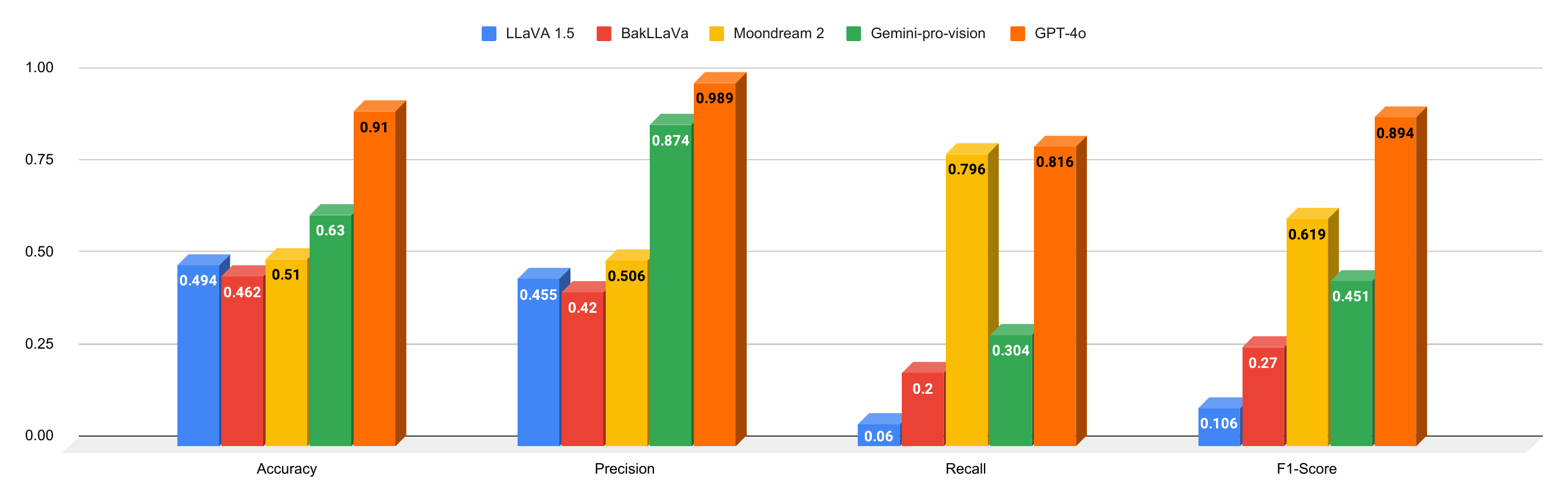}
        \caption{Performance of each LMM with prompt 2}
        \label{fig:prompt2}
    \end{minipage}\\ % Allows a small space between the images
    
    \begin{minipage}{0.99\textwidth}
        \centering
        \includegraphics[width=\textwidth]{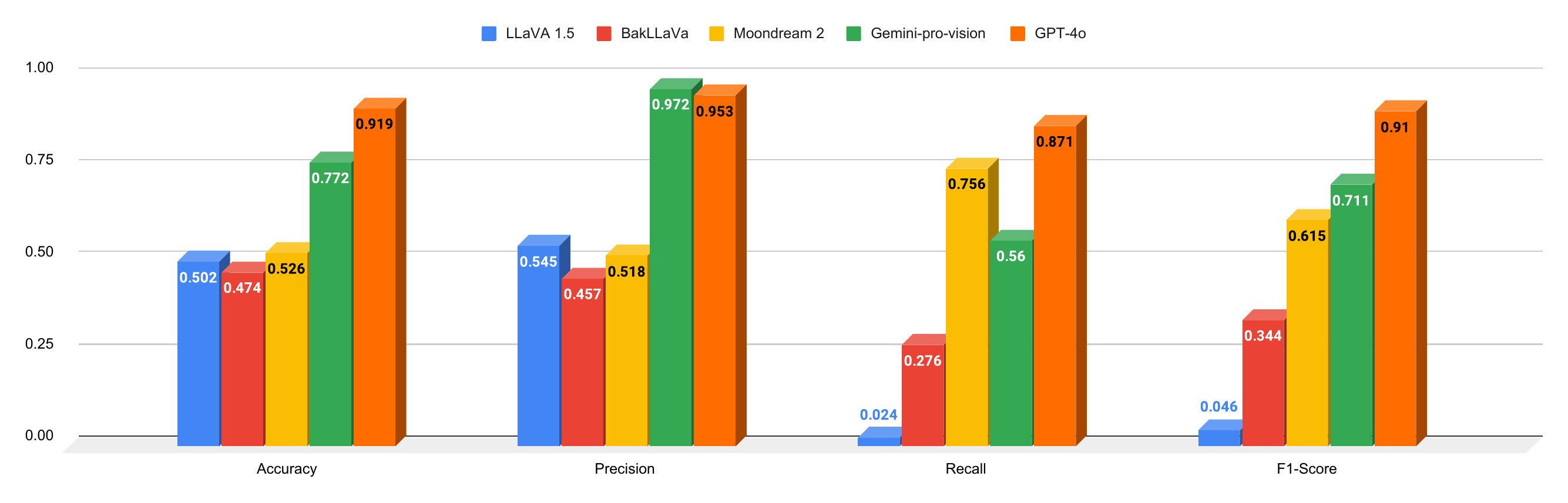}
        \caption{Performance of each LMM with prompt 3}
        \label{fig:prompt3}
    \end{minipage}
\end{figure}

\begin{itemize}
    \item \textbf{Prompt 1:} This baseline prompt tasked each LMM with classifying images as 'Triggered' or 'Unaltered' without additional details. Here, all models exhibit poor performance as shown in Figure \ref{fig:prompt1}.
    \begin{itemize}
        \item LLaVA showed modest accuracy at 51\%, but with a very low recall of 3.2\% and an F1-score of 6.1\%, indicating severe under-detection of triggered cases despite a precision of 72.7\%.
        \item BakLLaVa had a similar accuracy at 51.6\%, but exhibited a much higher recall of 66.4\%, balanced by a lower precision of 51.2\%, resulting in an F1-score of 57.8\%.
        \item Moondream 2 classified all images as triggered, achieving 50\% accuracy and precision, with a perfect recall of 100\%, resulting in an F1-score of 66.7\%.
        \item Gemini-pro-vision predicted all images as unaltered, which resulted in a 50\% accuracy and 0\% recall. The precision and F1-score could not be calculated here, because no instances were classified as triggered.
        \item GPT-4o had a low accuracy of 52.4\%, with extremely high recall of 98.3\% and a high precision, resulting in an F1-score of 65.9\%.
    \end{itemize}

    \item \textbf{Prompt 2:} This prompt included a specific description of the trigger as a small white square to guide the LMM on what to look for in the input image. As a result, the models that performed better are Gemini-pro-vision and GPT-4o, as shown in Figure \ref{fig:prompt2}:
    \begin{itemize}
        \item LLaVA's accuracy slightly decreased to 49.4\%, with low recall and precision, and an F1-score of 10.6\%.
        \item BakLLaVa showed a further decrease in accuracy to 46.2\%, with improved precision to 42\% and an F1-score of 27\%.
        \item Moondream 2 saw a slight accuracy increase to 51\%, with high recall (79.6\%) but moderate precision (50.6\%), resulting in an F1-score of 61.9\%.
        \item Gemini-pro-vision achieved a notable accuracy increase to 63\%, with precision jumping to 87.4\%, but a lower recall of 30.4\% suggested missed 'Triggered' cases, leading to an F1-score of 45.1\%.
        \item GPT-4o excelled under this prompt with an accuracy of 91\%, high precision (98.9\%), and substantial recall (81.6\%), resulting in an F1-score of 89.4\%.
    \end{itemize}
    \item \textbf{Prompt 3:} This prompt explicitly directed the models to examine the four corners of the image for a white square in an effort to direct the attention of the LMM to specific parts of the image. The results of applying this prompt to each LMM are shown in Figure \ref{fig:prompt3}:
    \begin{itemize}
        \item LLaVA 1.5's performance remained low with an accuracy of 50.2\%, and extremely low recall of 2.4\%, resulting in an F1-score of 4.6\%.
        \item BakLLaVa's accuracy was slightly better at 47.4\%, with a moderate increase in recall to 27.6\% and an F1-score of 34.4\%.
        \item Moondream 2 showed marginal improvement with accuracy at 52.6\% and high recall at 75.6\%, leading to an F1-score of 61.5\%.
        \item Gemini-pro-vision's accuracy soared to 77.2\%, with high precision (97.2\%) and more than half the cases correctly recalled (56\%), resulting in an F1-score of 71.1\%.
        \item GPT-4o again performed the best with an accuracy of 91.9\%, very high precision (95.3\%), and an 87.1\% recall rate, leading to an F1-score of 91\%.
    \end{itemize}
\end{itemize}

This detailed analysis reveals crucial insights into the strengths and limitations of prompt-engineered LMMs in image-based cybersecurity tasks. The results show that while tailored prompts can enhance performance significantly for some LMMs like Gemini-pro-vision and GPT-4o, other models still encounter challenges and show no clear improvement. 

\subsubsection{Fine-tuning ViT}
As outlined in the methodology, we fine-tune the ViT model using the training set and subsequently evaluate it on the same test set used for the prompt engineering strategies. The fine-tuning process resulted in optimal performance, achieving 100\% in accuracy, precision, recall, and F1-Score. This perfect score underscores the effectiveness of fine-tuned ViT models in tasks like trigger detection, where their ability to focus on specific regions of an image is crucial. Given the visually evident nature of the task, such a high level of performance is anticipated. The ViT's architectural strengths in image analysis are particularly well-suited to tasks where visual cues are clear and distinct, as evidenced by this outcome.

\subsection{Task 2: Malware Classification}
\subsubsection{Prompt engineering with LMMs}
\begin{figure}[htbp]
    \centering
    \includegraphics[width=0.99\textwidth]{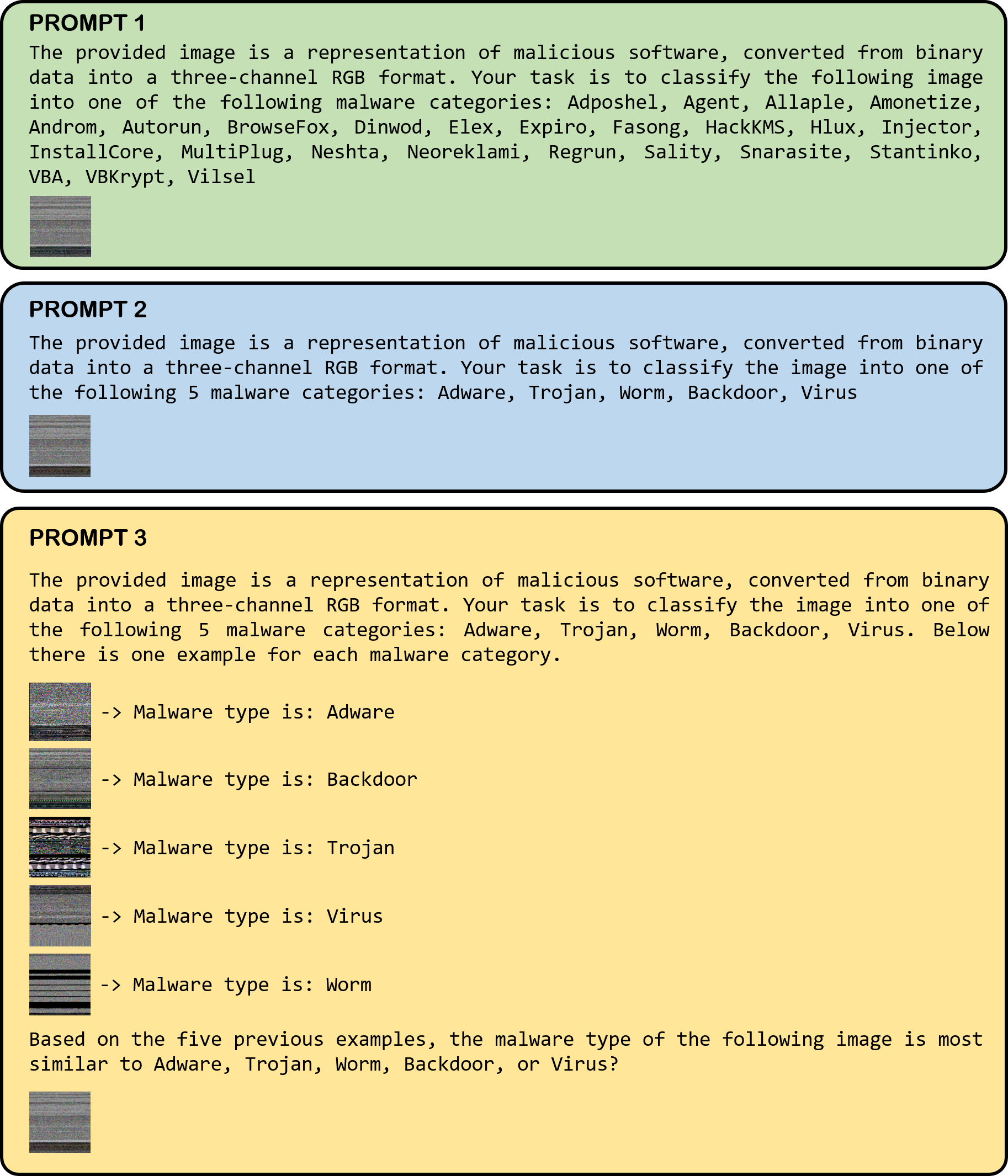}
    \caption{The three prompts used for malware classification.}
    \label{fig:malwareprompts}
\end{figure}

 Since in the visually-evident task, only Gemini-pro-vision and GPT-4o showed acceptable performance, these are the only models that will be used for the malware classification task since it is more complex. We use the flow outlined in the methodology section and apply it to the selected LMMs. We vary the input prompts to observe how this affects the results. Notably, we investigate three prompt templates to categorize malware images, as shown in Figure \ref{fig:malwareprompts}.

\begin{figure}[ht]
    \centering
    \begin{minipage}[b]{0.49\textwidth}
        \centering
        \includegraphics[width=\textwidth]{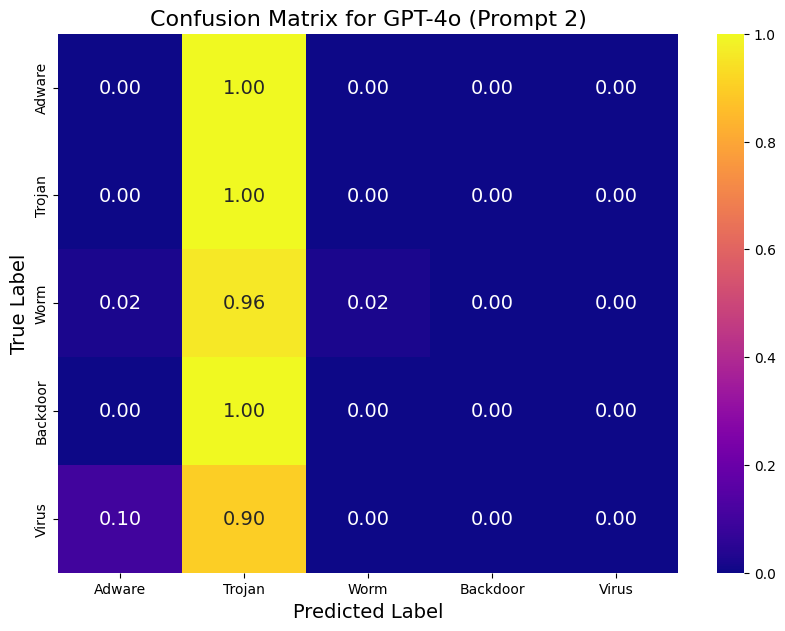}
        %\caption{Confusion Matrix for GPT-4o (Prompt 2)}
        \label{fig:gpt4o-prompt2}
    \end{minipage}\hfill
    \begin{minipage}[b]{0.49\textwidth}
        \centering
        \includegraphics[width=\textwidth]{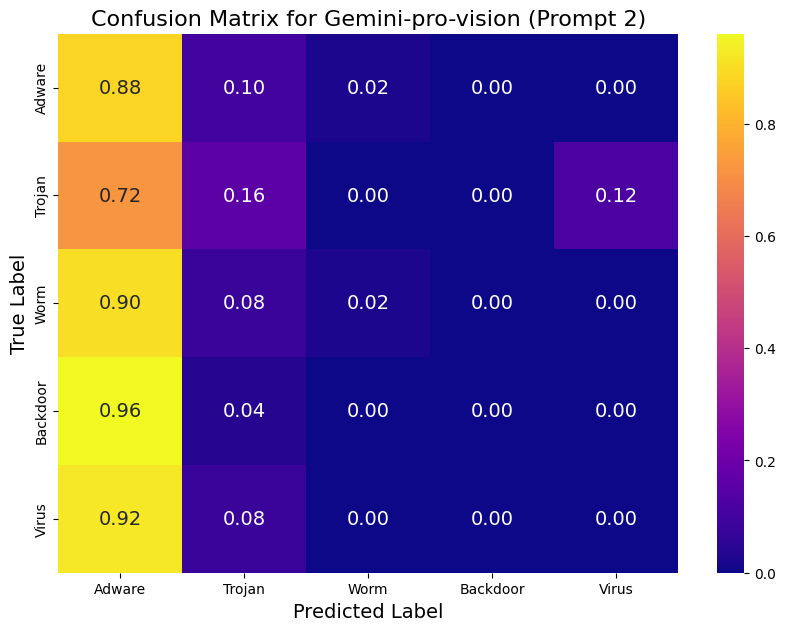}
        %\caption{Confusion Matrix for Gemini-pro-vision (Prompt 2)}
        \label{fig:gemini-prompt2}
    \end{minipage}
    \caption{Confusion matrices for Gemini-pro-vision and GPT-4o with prompt 2}
    \label{fig:malwareprompt2}
\end{figure}

\begin{figure}[ht]
    \centering
    \begin{minipage}[b]{0.49\textwidth}
        \centering
        \includegraphics[width=\textwidth]{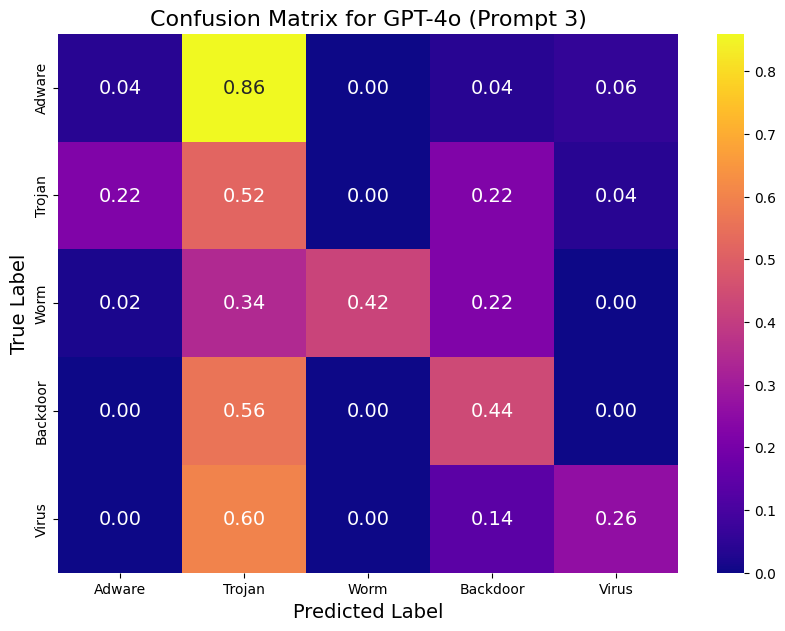}
        %\caption{Confusion Matrix for GPT-4o (Prompt 3)}
        \label{fig:gpt4o-prompt3}
    \end{minipage}\hfill
    \begin{minipage}[b]{0.49\textwidth}
        \centering
        \includegraphics[width=\textwidth]{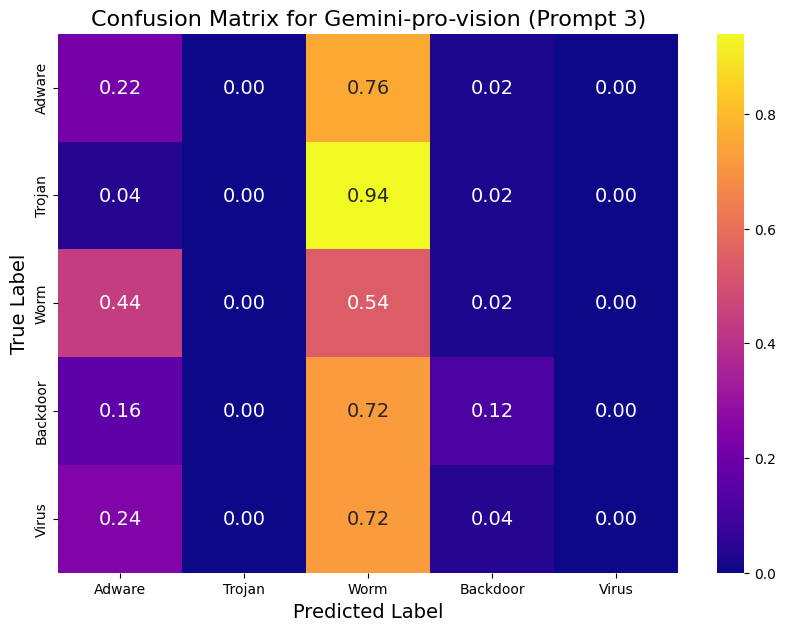}
        %\caption{Confusion Matrix for Gemini-pro-vision (Prompt 3)}
        \label{fig:gemini-prompt3}
    \end{minipage}
    \caption{Confusion matrices for Gemini-pro-vision and GPT-4o with prompt 3}
    \label{fig:malwareprompt3}
\end{figure}

\begin{itemize}
    \item \textbf{Prompt 1:} Starting with a baseline prompt, we ask the LMM to classify images into one of 25 malware types. The two models consistently identifiy images as belonging to the same class, irrespective of the input. Gemini-pro-vision always predicted "Fasong" as the malware class, and GPT-4o always predicted "Allaple" as the malware class. This behavior indicates a significant limitation in the LMM's ability to distinguish between various malware types. This challenge arises possibly due to the lack of highly specific and contextual information in the prompt, essential for such a fine-grained classification task.

    \item \textbf{Prompt 2:} To address the limitations observed in Prompt 1, we simplified the task by focusing on classifying into five malware families instead of 25 types. Despite this reduction in complexity, the models still exhibit a bias towards specific categories, although they occasionally predict other categories as well. The confusion matrices, as depicted in Figure \ref{fig:malwareprompt2}, show that even with simplified criteria, the LMMs struggle with accurate classification. Specifically, Gemini-pro-vision shows a pronounced bias towards the "Adware" class, while GPT-4o tends towards the "Trojan" class. The overall accuracies of Gemini-pro-vision and GPT-4o are 21.2\% and 20.4\%, respectively, with their macro F1-scores being 11.7\% and 7.61\%, respectively. This difficulty suggests that the task's inherent complexity and the LMMs' limitations in extracting sufficient context from prompts alone present significant challenges. Additionally, the nature of visual data in malware images, which may not be distinctly differentiable through prompts alone, could be contributing to these challenges.

    \item \textbf{Prompt 3:} In an attempt to improve classification accuracy, we introduce example images in the prompt for each malware family, asking the model to classify new images based on similarity. The confusion matrix in Figure \ref{fig:malwareprompt3} shows a marginal improvement in performance. However, there is still some bias towards a specific class: "Worm" in the case of Gemini-pro-vision and "Trojan" in the case of GPT-4o. Gemini-pro-vision achieves an accuracy of 17.6\% and a macro F1-score of 12.7\%, whereas GPT-4o achieves an accuracy of 33.6\% and a macro F1-score of 34.63\%. Even if GPT-4o shows more improvement compared to Gemini-pro-vision, the results still highlight the inherent difficulties faced by LMMs in processing and classifying complex visual data such as malware images. This suggests that while context and examples aid in performance, they are not sufficient to overcome the LMM's challenges in handling detailed and nuanced visual classifications.
\end{itemize}

\subsubsection{Fine-tuning ViT to Predict Malware Classes}
\begin{figure}[h]
    \centering
    \includegraphics[width=0.99\textwidth]{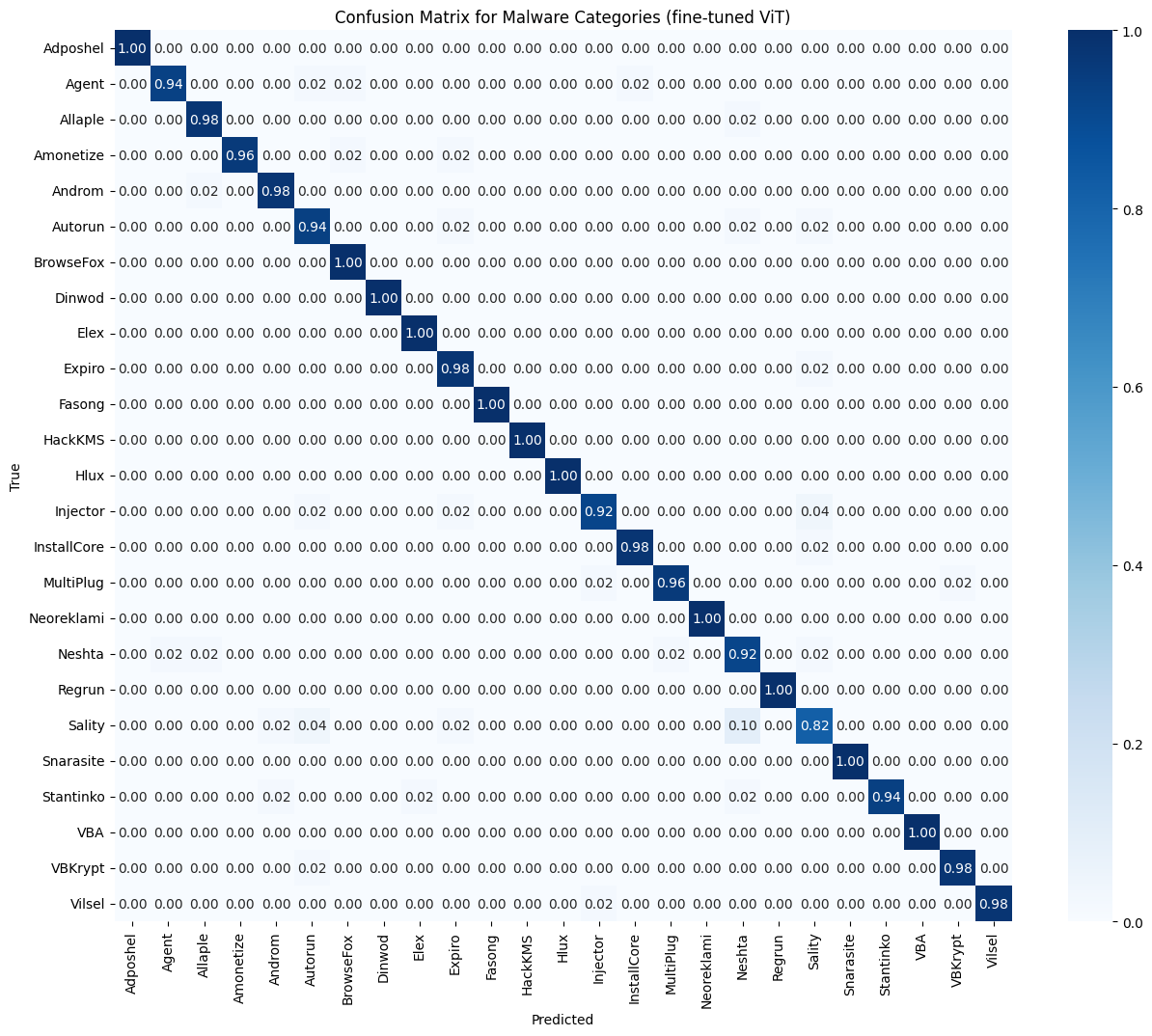}
    % Include figure here for ViT confusion matrix
    \caption{Confusion matrix for the fine-tuned ViT model in malware classification}
    \label{fig:vitmalwareclass}
\end{figure}

We fine-tune a ViT model to predict the 25 malware classes of the dataset. We report the results on the same test set we used to evaluate LMMs. The model achieved an accuracy of 97.12\%, a precision of 97.17\%, a recall of 97.11\%, and an F1-Score of 97.11\%. The high performance of the fine-tuned ViT model in accurately classifying malware classes is indicative of its robust pattern recognition capabilities. The confusion matrix shown in Figure \ref{fig:vitmalwareclass} reveals that the model is accurately predicting each of the 25 malware categories.
This result is particularly notable, considering the complexity and variety of malware signatures, highlighting the effectiveness of fine-tuning in enhancing model specificity for intricate classification tasks.

\subsubsection{Fine-tuning ViT to Predict Malware Families}
\begin{figure}[h]
    \centering
    \includegraphics[width=0.99\textwidth]{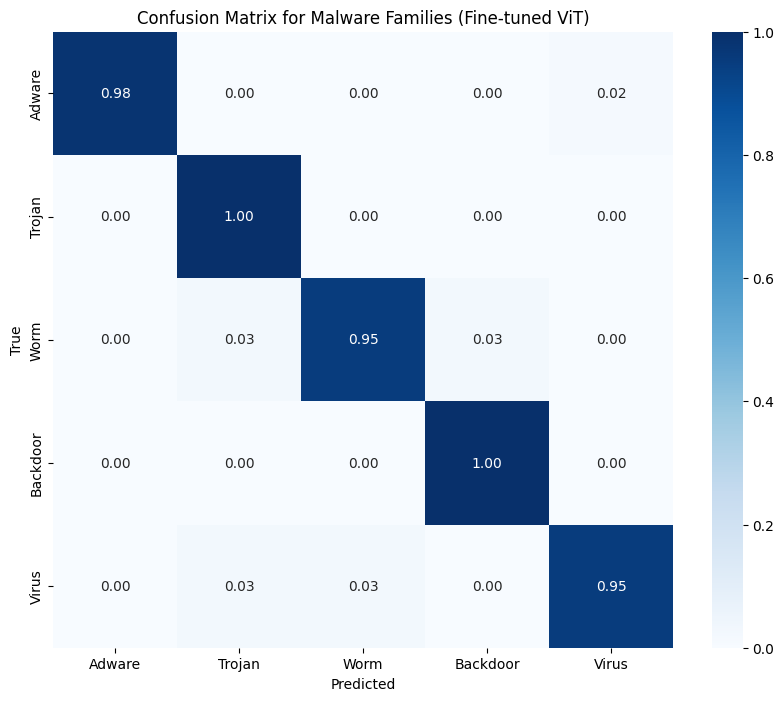}
    \caption{Confusion matrix for fine-tuned ViT model on malware family prediction}
    \label{fig:vitmalwarefamily}
\end{figure}

We fine-tune a ViT model to predict the 5 malware families of the dataset and report the results on the test set. The model achieved an accuracy of 98.00\%, a precision of 97.59\%,  a recall of 97.67\%, and an F1-Score of 97.61\%. The performance of the fine-tuned ViT in identifying malware families further establishes its superior capability in visual pattern analysis and classification. The confusion matrix shown in Figure \ref{fig:vitmalwarefamily} proves the model's ability to differentiate between multiple families with high accuracy. These results underscore the potential of ViT models in cybersecurity applications, where the ability to discern subtle differences in visual data is crucial for accurate threat detection and classification.

\section{Discussion}
The results from our experiments offer new insights into the capabilities and limitations of prompt-engineered LMMs and fine-tuned ViTs for cybersecurity applications. In this section, we perform a comparative analysis of both approaches across the two tasks, underlining the inherent challenges associated with both visually evident and visually non-evident tasks in the context of LMMs and ViTs.

The results from the trigger detection task reveal that even visually evident tasks can pose challenges for LMMs. While models like Gemini-pro-vision and GPT-4o  demonstrated improvement with more detailed prompting, as seen in the increased accuracy from Prompt 1 to Prompt 3, their performance still fell short of perfection, as evidenced by the results reported in Section 5. This indicates that the effectiveness of LMMs in visually evident tasks is highly contingent on the precision and context provided in the prompts. The inability of the model to achieve full accuracy, despite the triggers being visually evident, raises questions about the applicability of LMMs in scenarios where rapid and accurate detection of visual anomalies is critical. Moreover, through the experiments, it was demonstrated that even with prompt engineering, not all LMMs exhibit improvement. For instance, LLaVA, BakLLaVA, and moondream2 show almost a constant performance across the three prompts used.

On the other hand, our experiments in malware classification, a visually non-evident task, highlight a significant challenge for LMMs. The limited success in classifying malware types and families, even with enhanced prompts and example images, underscores the difficulty LMMs face in tasks requiring deep contextual understanding and intricate pattern recognition. As illustrated in Figures \ref{fig:malwareprompt2} and \ref{fig:malwareprompt3}, while prompt engineering offers some improvement, the models struggled to differentiate between various malware classes, indicating a gap in their ability to interpret complex visual data. This suggests that while LMMs can be powerful tools for certain types of image analysis, their effectiveness diminishes in tasks where visual cues are less obvious or require sophisticated interpretation.

In contrast, the fine-tuned ViTs demonstrated exceptional performance in both tasks. The ViTs achieved perfect scores in the visually evident trigger detection task and remarkable accuracy in the more complex malware classification task, as shown in Figures \ref{fig:vitmalwareclass} and \ref{fig:vitmalwarefamily}. These results underscore the fine-tuned ViTs' robustness in handling both visually evident and non-evident tasks, owing to their ability to focus on and analyze specific regions of an image. The success of fine-tuned ViTs in these experiments suggests their better suitability for cybersecurity applications, where the accuracy and reliability of visual analysis are paramount.

The findings suggest that while prompt engineering can enhance the utility of LMMs to some extent, the approach may not always be sufficient for tasks requiring high-level visual comprehension or pattern recognition. On the other hand, fine-tuned ViTs, with their advanced image processing capabilities, emerge as a more reliable and effective solution for diverse cybersecurity tasks, ranging from the detection of simple visual triggers to the classification of complex malware signatures.

In conclusion, our comparative analysis reveals that while LMMs are easier to use and offer a wide range of applicability, they do not perform well in all circumstances. This is particularly evident in tasks requiring high-level visual comprehension or intricate pattern recognition, where LMMs, even with prompt engineering, may fall short. As cybersecurity threats continue to evolve, the adoption of advanced vision models like ViTs, specifically fine-tuned for dedicated tasks, could prove instrumental in developing more robust and intelligent defense mechanisms against a wide array of cyber threats. The superior performance of ViTs in both visually evident and visually non-evident tasks highlights their potential as a reliable and effective solution for diverse cybersecurity challenges.

\section{Conclusion}
This study conducted a comparative analysis of prompt-engineered LMMs and fine-tuned ViT models in two distinct cybersecurity tasks: trigger detection (visually evident) and malware classification (visually non-evident). Employing a methodical approach that included prompt engineering for five LMMs and fine-tuning of pre-trained ViTs, the study aimed to assess the effectiveness of these models in specialized visual tasks.

The results demonstrated varying performance levels of LMMs and ViTs. LMMs, while user-friendly and adaptable to a range of applications, faced limitations in tasks requiring detailed visual comprehension, as seen in their performance in both trigger detection and malware classification. Specifically, LMMs showed improvement in trigger detection with more explicit prompting but did not achieve optimal accuracy. In the complex task of malware classification, LMMs struggled to accurately classify the diverse visual representations of malware.

In contrast, fine-tuned ViTs displayed a higher degree of effectiveness in both tasks. These models achieved perfect accuracy in trigger detection and notably high accuracy in malware classification, underscoring their capability to handle tasks that involve intricate visual analysis.

Based on the study's findings, future research could explore several areas:
\begin{itemize}
    \item Enhancing the efficiency and effectiveness of prompt engineering techniques for LMMs, especially in complex visual tasks.
    \item Expanding the application of fine-tuned ViTs across a broader range of cybersecurity challenges, including those involving more subtle and sophisticated visual patterns.
    \item Investigating the interpretability of these models, particularly ViTs, to increase transparency and trustworthiness in AI-driven cybersecurity solutions.
\end{itemize}

\section*{Acknowledgements}
The authors would like to acknowledge that this work has been supported by the Maroun Semaan Faculty of Engineering and Architecture (MSFEA) at the American University of Beirut (AUB)

%Bibliography
\bibliographystyle{unsrt}  
\bibliography{references}

\end{document}